\documentclass[10pt,journal,compsoc]{IEEEtran}
\usepackage{amssymb}
\usepackage[utf8]{inputenc}
\usepackage[english]{babel}
\usepackage{bbm}
\usepackage[normalem]{ulem}
\usepackage{graphicx}
\usepackage[inline]{enumitem}
\usepackage{tikz}
\useunder{\uline}{\ul}{}
\usepackage{amsmath}
\usepackage{caption}
\usepackage{natbib}
\setcitestyle{authoryear,open={(},close={)}}
\usepackage{url}
\usepackage{breakurl}
\usepackage[breaklinks]{hyperref}
\usepackage{breakcites}
\usepackage{microtype}
\usepackage{balance}
\newcommand*{\newtext}{}

\begin{document}

\title{A Review of Incident Prediction, Resource Allocation, and Dispatch Models for Emergency Management}


\author{Ayan~Mukhopadhyay, Geoffrey~Pettet, Sayyed~Mohsen~Vazirizade, Di~Lu, Alex~Jaimes, Said~El~Said, Hiba~Baroud, Yevgeniy~Vorobeychik, Mykel~Kochenderfer, Abhishek~Dubey
\IEEEcompsocitemizethanks{\IEEEcompsocthanksitem A. Mukhopadhyay was with Stanford University when this work was started.  He is currently with the Department of Electrical Engineering and Computer Science at Vanderbilt University\protect\\
E-mail: \href{mailto:ayan.mukhopadhyay@vanderbilt.edu}{ayan.mukhopadhyay@vanderbilt.edu}
\IEEEcompsocthanksitem G. Pettet, S. M. Vazirizade, H. Baroud, and A. Dubey are with Vanderbilt University
\IEEEcompsocthanksitem D. Lu and A. Jaimes are with DataMinr
\IEEEcompsocthanksitem S. Said is with Tennessee Department of Transportation
\IEEEcompsocthanksitem Y. Vorobeychik is with Washington University at St. Louis
\IEEEcompsocthanksitem M. Kochenderfer is with Stanford University}
}

\IEEEtitleabstractindextext{%
\begin{abstract}
In the last fifty years, researchers have developed statistical, data-driven, analytical, and algorithmic approaches for designing and improving emergency response management (ERM) systems. The problem has been noted as inherently difficult and constitutes spatio-temporal decision making under uncertainty, which has been addressed in the literature with varying assumptions and approaches. This survey provides a detailed review of these approaches, focusing on 
the key challenges and issues regarding four sub-processes: (a) incident prediction, (b) incident detection,
(c) resource allocation, and (c) computer-aided dispatch for emergency response.  We highlight the strengths and weaknesses of prior work in this domain and explore the similarities and differences between different modeling paradigms. We conclude by illustrating open challenges and opportunities for future research in this complex domain.
\end{abstract}
\begin{IEEEkeywords}
Resource Allocation for Smart Cities, Incident Prediction, Computer aided dispatch, Decision making under uncertainty
\end{IEEEkeywords}}
\maketitle

\begin{center}
To appear at Elsevier Accident Analysis \& Prevention
\end{center}

\section{Introduction}

Emergency response management (ERM) is a challenge faced by communities across the globe.
First responders need to respond to a variety of incidents such as fires, traffic accidents, and medical emergencies. 
They must respond quickly to incidents to minimize the risk to human life~\citep{jaldell2017important,jaldell2014time}. Consequently, considerable attention in the last several decades has been devoted to studying emergency incidents and response.
Data-driven models help reduce both human and financial loss as well as improve design codes, traffic regulations, and safety measures. Such models are increasingly being adopted by government agencies. Nevertheless, emergency incidents still cause thousands of deaths and injuries and result in losses worth billions of dollars directly or indirectly each year~\citep{crimeUS}. This is in part due to the fact that emergency incidents (like accidents, for example) are on the rise with rapid urbanization and increasing traffic volume.  

\begin{figure*}[t]
\centering
\begin{center}
\includegraphics[width=0.9\textwidth]{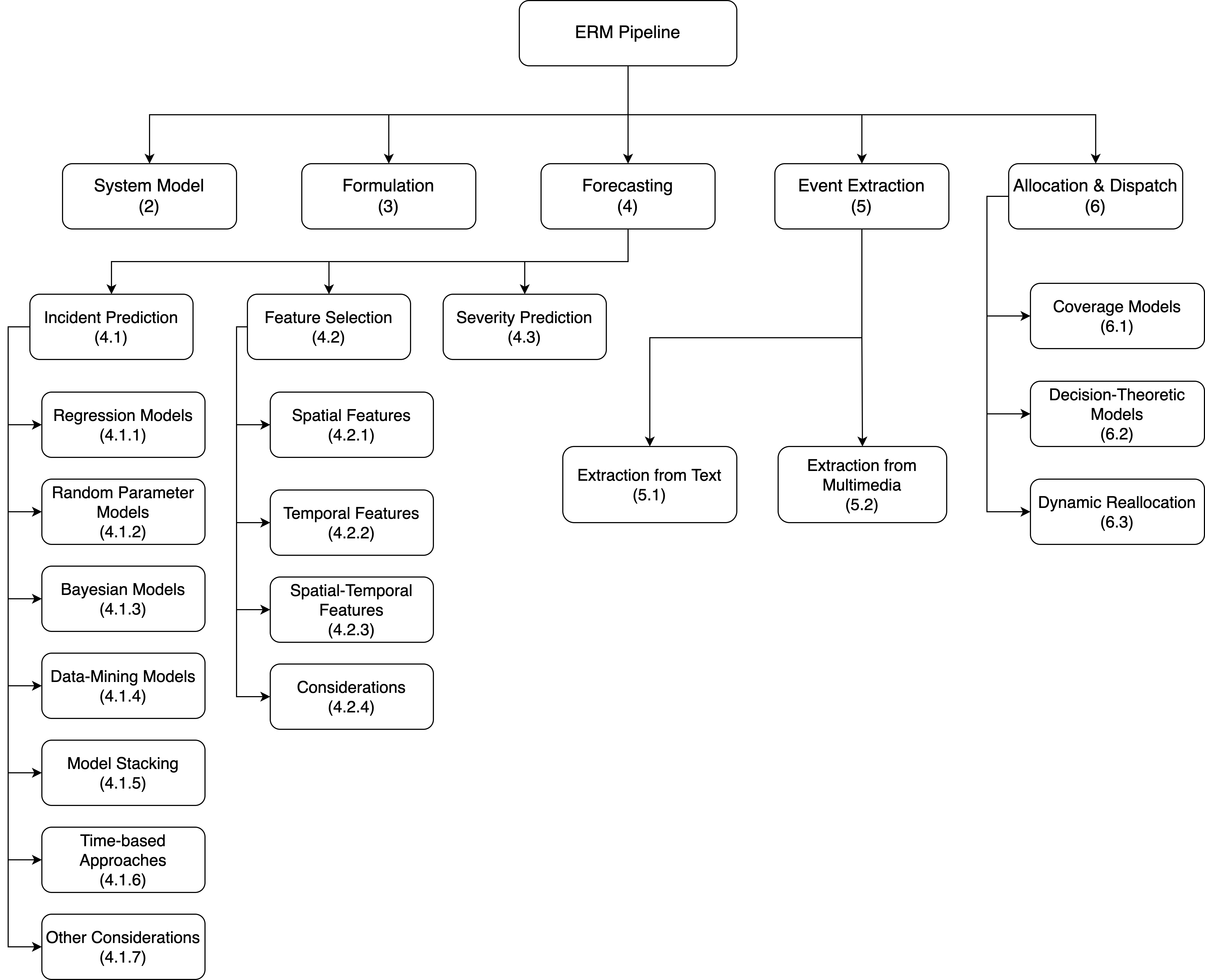}
\caption{Outline for the survey: Numbers in each box link the topic to a specific section in this paper. 
}
\label{fig:outline}
\end{center}
\end{figure*}

ERM can be divided into five major components: 1) mitigation, 2) preparedness, 3) detection, 4) response, and 5) recovery. While most prior work has identified mitigation, preparedness, response, and recovery as the primary components of ERM systems~\citep{MukhopadhyayDissertation2019,dhs},
crowd-sourced information and additional sensors have motivated deployment of technology to provide early detection of incidents (before someone calls for help).  Mitigation involves sustained and continuous efforts to ensure safety and reduce long-term risks to people and property. It also involves understanding \textit{where} and \textit{when} incidents occur and designing predictive models for both risk and incident occurrence. Preparedness involves creating infrastructure that enables emergency response management. This stage involves selecting stations for housing responders, ambulances, and police vehicles as well as designing plans for response. The third phase seeks to use automated techniques to detect incidents as they happen in order to expedite response. The fourth phase, arguably the most crucial, involves dispatching responders when incidents happen or are about to occur. Finally, the recovery phase ensures that impacted individuals and the broader community can cope with the effects of incidents. 
While most prior work in ERM has studied these problems independently, these stages are actually inter-linked, with the output of one stage serving as input for another. For example, predictive models learned in the \textit{preparedness} stage are used in planning \textit{response} strategies. Therefore, it is crucial that ERM pipelines are designed in a manner that considers such intricate inter-dependencies.

\begin{figure*}[t]
\centering
\begin{center}
\includegraphics[height=3.5cm]{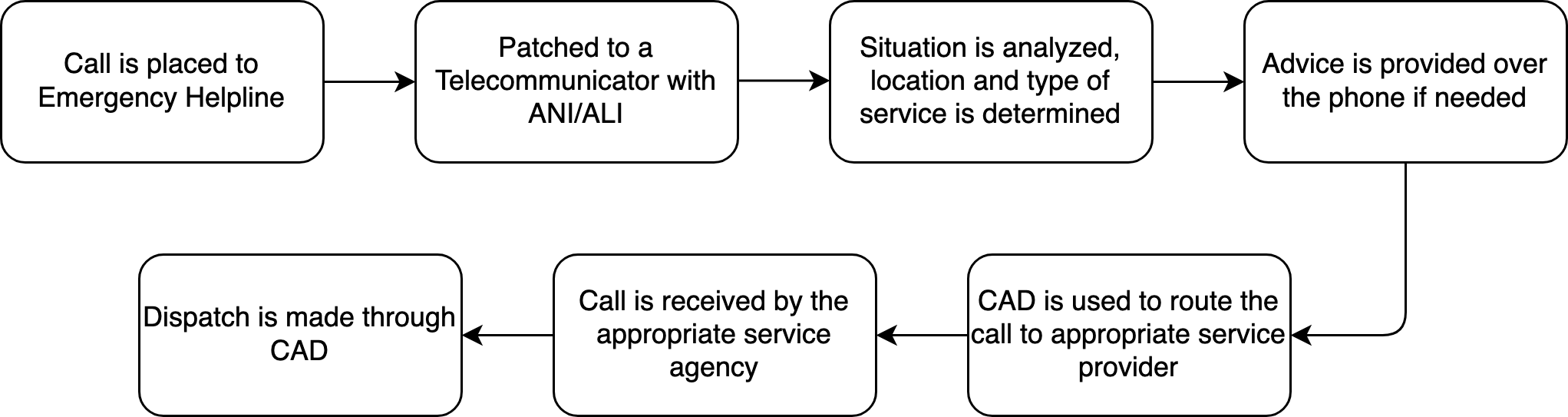}
\caption{Typical Emergency Dispatch Helpline Model: first responders analyze the type of call, offer immediate help and use computer-aided dispatch to allocate specific resources to incidents.}
\label{fig:911model}
\end{center}
\end{figure*}

In this survey, we cover prior work on some of the most widely explored approaches that fall into the categories of mitigation, preparedness, detection, and response, and explain how the overall ERM pipeline functions. \newtext{First, we address the scope of the problem and precisely define the incidents that we consider in the paper.} One way to categorize incidents is by the rate at which they occur and how they affect first responders. For example, 
some incidents happen often, and addressing them is part of the day-to-day operations of first-responders. Examples of such incidents include crimes, accidents, calls for medical services, and urban fires. A second category consists of comparatively less frequent incidents, which include natural calamities like earthquakes, floods, and cyclones. 
\newtext{To scope our research we primarily focus on principled approaches to address frequent urban incidents like road accidents. Having a narrow focus regarding the type of the incident enables us to explore the large spectrum work dedicated to designing principled approaches to ERM.} Our primary reason to focus on urban emergency incidents is simply the alarming extent of the damage such incidents cause and the sheer frequency of their occurrence. Globally, about 3,200 people die every day from road accidents alone, leading to a total of 1.25 million deaths annually~\citep{roadStats}. In fact, it is noted that without appropriate measures, road accidents are set to be the fifth largest cause of death worldwide by 2030~\citep{asirt}. Calls for emergency medical services (EMS) are also a major engagement for first responders, and there are more than 240 million EMS calls made annually in the United States alone~\citep{911Stats}. Therefore, it is imperative that we design principled approaches to understand the spatial and temporal characteristics of such incidents and investigate algorithmic methods that can mitigate their effects.

In this survey, we explore models and approaches to design principled approaches to ERM from various fields like operations research, transportation engineering, statistics, and machine learning in order to understand commonalities and differences among them. We seek to provide a unified perspective on such systems. There are comprehensive reviews on crash prediction models~\citep{nambussiReview,yannis2017road,kiattikomol2005freeway}, emergency facility location approaches~\citep{li2011covering}, and dispatch strategies~\citep{bohm2018accuracy}. \newtext{In particular, the work by \citeauthor{Lord2010}~\citep{Lord2010} provides a  particularly insightful summary of crash prediction models.} However, to the best of our knowledge, there is no comprehensive study that links prediction models from different persepectives and investigates covariates of relevance, modeling paradigms, and planning approaches comprehensively. We treat the ERM system in its entirety and provide a comprehensive survey of prior work done in the fields of predictive modeling, event extraction, and planning approaches to aid emergency response. This survey provides a framework for future research on integrated emergency response management pipelines for smart and connected communities.

We show a brief outline of the survey in Figure~\ref{fig:outline}. We begin by providing an overview of the system model for ERM pipelines in section \ref{sec:systemModel} and mathematical formulations for each of the sub-problems in section \ref{sec:formulation}. Then, we discuss approaches to incident prediction (section \ref{sec:IncidentAnalysis}), event extraction (section \ref{sec:extraction}), and allocation and dispatch (section \ref{sec:ResponderAllocationandDispatch}). We highlight key takeaways for practitioners at the end of each section. Finally, we identify key challenges, knowledge gaps, and opportunities for future work in section \ref{sec:future}.



\section{Understanding Emergency Response Systems}\label{sec:systemModel}

\begin{figure*}[t]
\begin{center}
\includegraphics[width=0.9\textwidth]{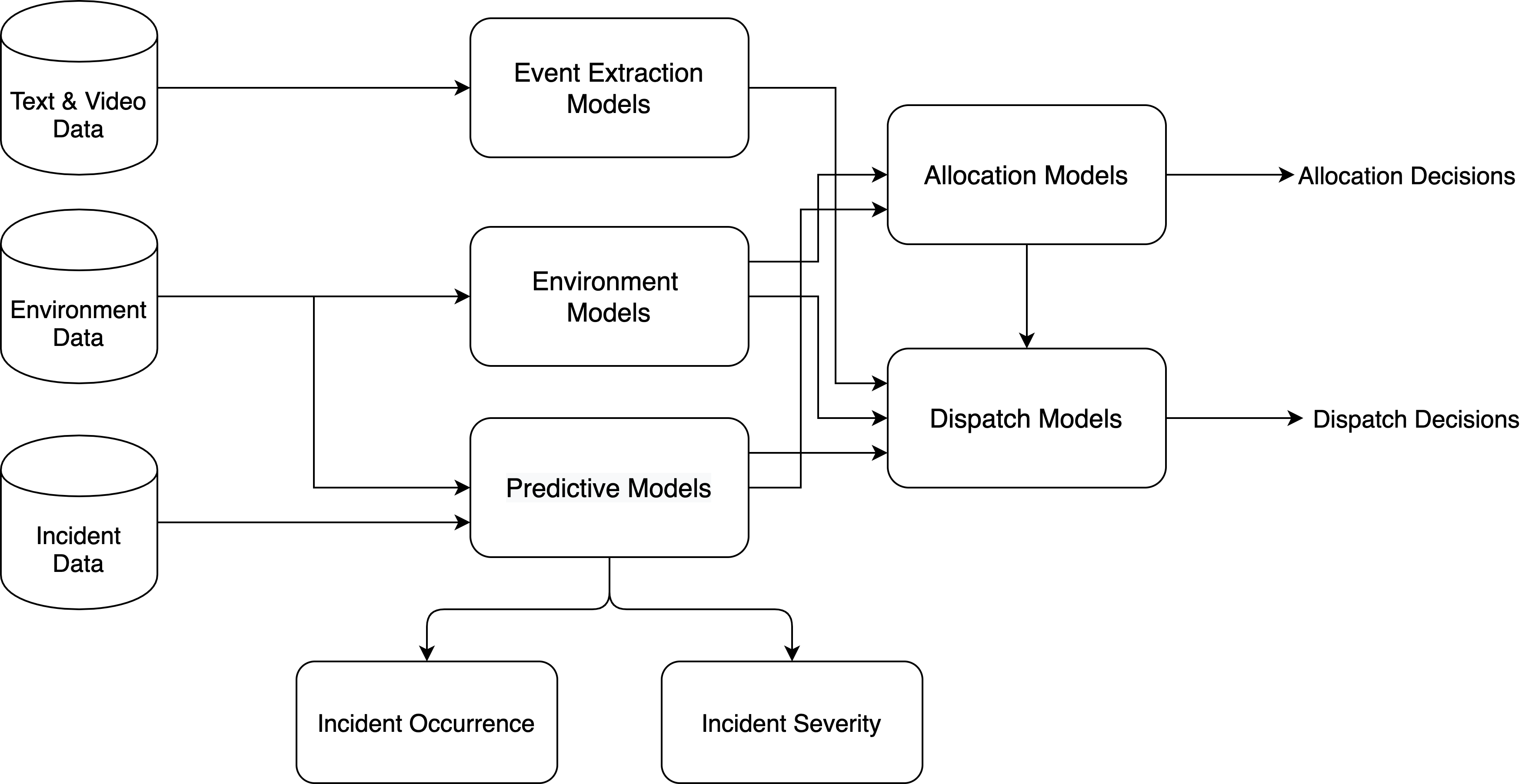}
\caption{ERM System Pipeline: historical data from different sources are used to design predictive models for incidents and the environment, which in turn are used to create allocation and response models. Events can be extracted using text and video data to expedite reporting and aid response.}
\label{fig:Sysmodel}
\end{center}
\end{figure*}

We study the problem of optimally responding to emergency incidents in urban areas. Incidents are reported to central emergency response agencies, which have streamlined mechanisms for processing the request. For example, in the United States, emergency helpline calls are placed by dialing 911. Calls can be made by many \textit{agents}. People in need of assistance or other people who might have observed an incident can report it to the concerned authorities. Such a report is typically referred to as a ``call'' for emergency services. It is also possible for first responders to automatically extract reports about incidents through social media or video feeds. We show the steps that follow a call for assistance in Figure~\ref{fig:911model}~\citep{911works}. The call is appended with automatic name and location information (ANI/ALI) and patched to a trained telecommunicator. The telecommunicator analyzes the situation and the type of response needed (EMS or fire, for example). In some cases, such as those requiring cardiopulmonary resuscitation (CPR), guidance might be provided through the phone before first-responders reach the scene. The call is then transferred to the concerned agency (such as the fire department) by a computerized mechanism. The agency then uses its computer-aided dispatch (CAD) system to dispatch a responder to the scene. This set of events defines an ERM system, and it governs the pipeline of incident response, including detecting and reporting incidents, monitoring and controlling a fleet of response vehicles, and finally dispatching responders when incidents occur. In many cases there are multiple organizations governing this pipeline for an urban area; for example, ambulances and police cars might be dispatched from different departments.

Agents\footnote{We use the term ``agents'' as is common in multi-agent systems community.} which respond to incidents like accidents and fires include ambulances, police vehicles, and fire trucks (among others) and are referred to as \textit{responders}. Responders are typically equipped with devices that facilitate communication to and from central control stations. In many cases, especially in the United States, responders like ambulances are equipped with computational devices like laptops as well. Once an incident is reported, responders are dispatched by a human agent to the scene of the incident (guided by some algorithmic approach like a CAD system). This process typically takes a few seconds,\footnote{This is based on our communication with fire departments in the United States ~\citep{fireDepartmentCommunication}; time taken to dispatch responders presumably varies across the globe.} but can be longer if dispatchers are busy. If no responder is available, the incident typically enters a waiting queue and is attended once a responder becomes free. Each responder is located in a specific \textit{depot} (fire-station, for example), which are situated at various points in the spatial area under consideration. Once a responder has finished servicing an incident, it is directed back to the depot and becomes available to be re-dispatched by the dispatcher while en-route. An aspect that plays a key role in dispatch algorithms is that if there are any free responders available when an incident is reported, one must be dispatched to attend to the incident. This constraint is a direct consequence of the bounds within which emergency responders operate, as well as the critical nature of the incidents.



The components of ERM that we focus on are shown in Figure~\ref{fig:Sysmodel}. ERM pipelines typically use data from historical incidents and the environment, including weather, road geometry, traffic patterns, and socio-economic data. It is also possible to use textual and video data to extract information about the occurrence of incidents. We divide an ERM system into five major components: 1) predictive models for incident occurrence, 2) event extraction models to detect incidents, 3) models for environmental features like traffic and weather, 4) allocation models to optimize the spatial locations of responders and depots, and 5) dispatch models to create algorithmic approaches to respond to incidents when they occur. These components are intricately linked, and the performance of each plays a crucial role in the overall performance of the ERM pipeline.

Incident prediction models form the basis of an ERM system. In order to mitigate the effects of incidents, it is important to understand \textit{where} and \textit{when} such incidents occur. Incident models are typically designed using historical incident data, but such models often use historical environmental data as well; for example, it is common for accident prediction models to use historical traffic data. Allocation models are then used to allocate responders in time and space in anticipation of future incidents. Finally, allocation and prediction models are used to create dispatch models, which can be thought of as a policy that guides real-time response. A rather recent trend in designing ERM systems is to include event extraction models in the pipeline, which can automate the discovery and reporting of incidents.

Significant prior work has focused on understanding and designing algorithmic approaches for each of the modular components. This article studies models for incident prediction, allocation, extraction, and dispatch. While we do not discuss models of relevant environmental factors, they are important to the development of the overall pipeline.

We focus this survey primarily on roadway accidents. The reason for this choice is two-fold. First, most prior work in incident analysis has focused on accidents and crashes, and this presents a rich body of work to survey and draw inferences about. Secondly, prediction, allocation and response to accidents and EMS calls involve similar characteristics and constraints from an algorithmic perspective. Both require dispatching of ambulances as quickly as possible, and from the scene of the incident, injured victims need to be transported to nearby hospitals. Therefore, most of our discussion on accidents can be broadened to EMS calls in general, but focusing on one particular type allows us to discuss various technical approaches in greater detail. 
\section{General Mathematical Formulation of the Decision Problem}\label{sec:formulation}
To help provide common context, we start by defining a broad mathematical formulation for incident prediction, extraction, and planning problems that we use throughout this survey. Given a spatial area of interest $S$,
the decision-maker observes a set of samples (possibly noisy) drawn from an incident arrival distribution. These samples are denoted by
$\{(s_1,t_1,k_1,w_1),(s_2,t_2,k_2,w_2),\dots,(s_n,t_n,k_n,w_n)\}$, where $s_i$, $t_i$, and $k_i$ denote the location, time of occurrence, and reported severity of the $i$th incident, respectively, and $w_i \in \mathbbm{R}^m$ represents a vector of features associated with the incident. We refer to this tuple of vectors as $D$, which denotes the input data that the decision-maker has access to. The vector $w$ can contain spatial, temporal, or spatio-temporal features and it captures covariates that potentially affect incident occurrence. For example, $w$ typically includes features such as weather, traffic volume, and time of day. The most general form of incident prediction can then be stated as learning the parameters $\theta$ of a function over a random variable $X$ conditioned on $w$. We denote this function by $f(X \mid w,\theta)$. The random variable $X$ represents a measure of incident occurrence such as a \textit{count} of incidents (the number of incidents in $S$ during a specific time period) or \textit{time} between successive incidents. The decision-maker seeks to find the \textit{optimal} parameters $\theta^*$ that best describe $D$. This can be formulated as a maximum likelihood estimation (MLE) problem or an equivalent empirical risk minimization (ERM) problem.

We review prior work focused on modeling the function $f(X \mid w,\theta)$. There have been many different approaches for modeling $f$. It can be modeled as an explicit probability density or mass (e.g., Poisson distribution), or a function that does not strictly conform to such definitions (e.g., a linear regression approach with $X$  being the dependent variable). Nonetheless, such functions typically have probabilistic interpretations, and we present different approaches of modeling $f$ in section \ref{sec:IncidentAnalysis}. We first highlight different modeling choices for understanding the spatial-temporal nature of accidents. Then, we focus on the vector $w$. Arguably, the most crucial part in learning a model over incident occurrence involves choosing $w$, and we review various covariates in section \ref{sec:FeatureSelection}. 


The next step in an emergency response pipeline is planning in anticipation of incidents. This involves stationing responders strategically and dispatching them as incidents occur. This process can be broadly represented by the optimization problem $\max_{y} G(y \mid f)$, where $y$ represents the decision variable (which typically denotes the location of emergency responders in space), $G$ is a reward function chosen by the decision-maker, and $f$ is the model of incident occurrence. For example, $G$ might measure the total coverage (spatial spread) of the responders, or the expected response time to incidents. Therefore, given $f$, the decision-maker seeks to maximize the function $G$. 


There are two major approaches for modeling the response problem. First, the planning problem can be represented as a stochastic control process. For example, it can be formulated as a Markov decision process (MDP) \citep{kochenderfer2015decision}. This formulation is particularly relevant for problems seeking to find policies for dispatch. The aim is to find an optimal policy (i.e. control choices for every possible state of the system) that maximizes the expected sum of rewards. The second approach is to directly model the planning problem as an optimization problem according to a specific measure of interest. As an example, a lot of prior work has focused on maximizing the coverage of emergency responders~\citep{toregas_location_1971,church1974maximal, gendreau_solving_1997}. 


It is important to note that emergency incidents can only be responded to after the concerned authorities are notified. However, often times in practice, there is a gap in time between the occurrence of an incident and the time at which a call is made for response. Event extraction models seek to bridge this gap by identifying incidents before they are reported to the responders. Mathematically, this process involves learning a function $E$ which takes relevant data $u$ as input (for example, text data from social media), and outputs information about an event (for example, the location of an accident). This can be represented as $\bar{z} = E(u)$, where $\bar{z}$ denotes an estimate of variable $z$. In event extraction, $z$ can denote any variable related to the incident such as the location of the incident, the extent of the damage, the number of people involved, and so forth.

\section{Incident Forecasting}\label{sec:IncidentAnalysis}
Incident forecasting is necessary to understand the likely demand of the emergency resources in a given region, and forms the basis for approaches to stationing and dispatch. We divide the discussion on incident forecasting into three major parts: 1) approaches to incident prediction, 2) features used in incident prediction and feature selection, and 3) predicting incident severity.

\subsection{Approaches to Incident Prediction}

Prior work has involved learning spatial-temporal models of incident occurrence. From our definition of incident prediction models described in section \ref{sec:formulation}, forecasting models correspond to determining the function $f$. An important method in incident prediction is known as `crash frequency analysis', which uses the frequency of incidents in a specific discretized spatial area as a measure of the inherent risk the area possesses~\citep{deacon1974identification}. \citeauthor{deacon1974identification}~\citep{deacon1974identification} identified key questions that practitioners should answer while designing predictive models for incident occurrence, and their work is still relevant to decision-makers and policy designers. This approach also forms the basis of \textit{hotspot} analysis~\citep{cheng2005experimental,nij2005}, which is widely used today as a relatively simple and fast method to visualize incident data. A shortcoming of frequency analysis is that it neglects fluctuations in incident occurrence and requires a large volume of incident data to infer accurate characteristics of occurrence~\citep{yu2014comparative,ryder2019spatial}. The core idea behind frequency analysis continues to be in use today; although it is common to use it in conjunction with other covariates of relevance and frame the overall problem as a regression model.


\subsubsection{Regression Models}
 
One of the earliest regression models used to model incident occurrence involved multiple linear regression models with Gaussian errors~\citep{frantzeskakis1994interurban,jovanis1986modeling}. However, modeling accident count by linear regression can be inaccurate, as the response variable is discrete and strictly positive. In addition, it has also been shown that linear regression models fail to model the sporadic nature of emergency incidents~\citep{miaou1993modeling,joshua1990estimating}. Linear regression models with multiplicative effects have also been investigated but have shown to be inaccurate compared to other models~\citep{miaou1993modeling}. 
The inaccuracies of linear regression methods in the context of accident prediction is investigated and summarized by \citeauthor{miaou1993modeling}~\citep{miaou1993modeling}. \citeauthor{rakha2010linear}~\citep{rakha2010linear} revisited this problem recently, and used data aggregation techniques to satisfy assumptions made by linear regression. While such an approach has shown performance on par with other regression models (Poisson regression, for example), it needs further validation before it is widely adopted.

The inaccuracies of linear regression and the suitability of Poisson models for count data led to the widespread use of Poisson regression for modeling incident occurrence~\citep{jovanis1986modeling}. Each incident is considered a result of an independent Bernoulli trial. Given that all the trials are generated by the same stochastic process, the series of trials can be modeled by a binomial distribution. As the number of trials becomes large and the probability of success is very small, the probability distribution over the count of incidents takes the form of a Poisson distribution~\citep{lord2005poisson}. To accommodate the feature vector $w$, Poisson regression assumes that the logarithm of the expected value of the distribution is a linear combination of $w$. This methodology has been used extensively for emergency incident analysis~\citep{bonneson1993estimation,maher1996comprehensive,sayed1999accident,joshua1990estimating,miaou1993modeling}.

An issue with using Poisson regression is that the expected value of the response variable (count of incidents) equals its variance. This is typically not the case with crash data, which is over-dispersed, meaning that the variance of the data is greater than its mean~\citep{lord2005poisson}. There are examples of incident data being under-dispersed as well~\citep{ye2018crash}. Therefore, the broader argument against the use of Poisson regression is that it might not be able to model real-world crash data, which can be under-dispersed or over-dispersed.
An approach to accommodate over-dispersion is to use Poisson-hierarchical models~\citep{Deublein2013}. Poisson-hierarchical models (as well as Poisson models) fall under the broader category of generalized-linear models (GLM), which is a family of distributions used widely in statistics and machine learning. From this family, the Poisson-gamma (also called negative binomial) and Poisson-lognormal models are particularly relevant. 

The Poisson-gamma model is a Poisson distribution whose mean parameter follows a gamma distribution. It has been shown that the Poisson-gamma model fits crash data better than Poisson models, and it has been extensively used for crash prediction~\citep{Quddus2008,akinCrashBinomial,ladron2004forecasting,caliendo2007crash,ackaah2011crash,dissanayake2006statistical}. While the Poisson-gamma model solves the problem of over-dispersion, it performs poorly on under-dispersed data and is particularly problematic to use with small sample sizes and with data with low sample mean~\citep{lord2008effects,aguero2008analysis}. The Poisson-lognormal model is conceptually the same as Poisson-gamma model, but it uses the lognormal distribution for the mean parameter rather than the gamma distribution~\citep{Shirazi2019, Park2007, Ma2008, Aguero-ValverdeJonathan2013FBPg}. The lognormal distribution is a heavy tail distribution and provides more flexibility for over-dispersion. Recently, the Poisson-inverse-gamma model has been used in crash modeling~\citep{Khazraee2018}. However, such models do not have closed-form MLE solutions unlike the Poisson-gamma models~\citep{Lord2010}.

Despite the success of Poisson and Poisson-hierarchical models, a common shortcoming is that both models fail to adequately handle the prevalence of zero counts in crash data~\citep{lord2005poisson}. A remedy to this problem is to use zero-inflated models, and both zero-inflated Poisson and zero-inflated Poisson-gamma models have been used to model accident data~\citep{qin2004selecting,lord2007further,huang2010modeling}. Zero-inflated models can be described as having dual states, one of which is the \textit{normal} state, and the other the \textit{zero} state. The excess zeros that cannot be explained by standard count-based models can then be considered to have arisen due to the presence of a separate state. Zero-inflated models result in improved statistical fit to accident data. However, \citeauthor{lord2005poisson}~\citep{lord2005poisson} 
note that most prior works justify the use of zero-inflated models by improved likelihood, and therefore automatically assume that crash data is generated by a dual-state process (except work by \citeauthor{miaou1993modeling}~\citep{miaou1993modeling}, which uses a zero-inflated model to justify misreporting of incidents). 
Through empirical data and simulations, they show that excess zeros could arise due to various other factors like low traffic exposure and the choice of spatial and temporal scales by the model designer. As a result, it is not clear if the statistical backing to using dual-state models is accurate or not. In our opinion, the work by \citeauthor{lord2005poisson}~\citep{lord2005poisson} is particularly profound, and the argument that statistical fit should not be the only consideration for fitting models to crash data (and other data in general) is extremely cogent. 

\subsubsection{Random-Parameter Models}\label{sec:pred:models:rp}
Accounting for unobserved heterogeneity (i.e., factors affecting incident frequency but not captured in the data) \newtext{has dominated recent statistical modeling development, with random-parameter (RP) models being among the most widely used approaches~\citep{mannering2016unobserved}.} \newtext{Unobserved heterogeneity introduces a variation in the effect of observed variables on the outcome. The outcome is typically the likelihood and severity of a crash. For example, a highway's design speed limit is a commonly used variable in the prediction of the likelihood of crashes. However, this may introduce unobserved heterogeneity if the vehicle's actual speed is not considered which may be different than the design speed limit across different drivers. Environmental conditions are also commonly used to explain crash occurrence and severity such as time of the day and weather variables. However, the same amount of precipitation may lead to different outcomes in the likelihood and severity of accidents depending on the geographic area and the different ways drivers respond to adverse conditions. Additionally, unobserved heterogeneity can result from the spatial or temporal aggregation of accidents. Since these events are rare, they are often aggregated over time (e.g., number of accidents per 4 hours) or space (e.g., number of accidents per road segment) before they are modeled. The lack of consideration for unobserved heterogeneity will lead to biased estimates because the effect of an observed variable will be the same across all observations for a particular instance \citep{mannering2016unobserved}. RP models address heterogeneity by allowing the estimated parameters to vary across observation according to a continuous distribution.} A significant portion of RP models in the literature are based on the assumption that random parameters follow a distribution with a common mean and no mutual dependence~\citep{el2009accident,milton2008highway}. However, lack of consideration of cross-correlation and mutual dependence can lead to biases in the estimation of parameter variances~\citep{conway1991important}.

A few recent studies have considered cross-correlated RP models and compared their performance to fixed-parameters and uncorrelated RP models. The correlated RP negative binomial model resulted in an improved log-likelihood compared to the fixed-parameters model~\citep{venkataraman2011model} and better statistical performance and predictive power compared to the uncorrelated model~\citep{coruh2015accident}. In another study, correlated RP Tobit model was shown to outperform both fixed-parameters and uncorrelated RP Tobit models~\citep{yu2015correlated}. However, these results are still not conclusive as other studies have found the relative statistical performance between uncorrelated and correlated RP count models to be comparable~\citep{saeed2019analyzing}. Therefore, additional research is needed to determine the advantages of correlated RP models. In addition to cross-correlations and improved statistical performance, another advantage of using correlated RP models is the ability to account for the heterogeneous effects of covariates across roadway segments as they apply to crash frequency analysis on multilane highways~\citep{saeed2019analyzing}. \newtext{While the focus of this section is on RP models as they are the most adopted methods, it is worth noting that other approaches have been developed to address unobserved heterogeneity (see the work by~\cite{mannering2016unobserved} for an extensive review). For instance, latent-class (finite mixture) models seek to identify groups of observations having homogeneous variable effects~\citep{cerwick2014comparison}. These models do not require a parametric assumption for the distribution of estimated parameters like RP models; however, they still impose a parametric model structure and can be computationally intensive. To account for the variation at both the group and individual observation levels, RP models within each class have been used with mixture models~\citep{xiong2013heterogeneous}. Other approaches address specific heterogeneity issues such as Markov-switching models which have been used for time-dependent unobserved heterogeneity~\citep{xiong2014analysis}. Such a form of heterogeneity can be caused by time-varying factors such as traffic and weather conditions or when the accidents are aggregated over a certain time period.}

\subsubsection{Bayesian Approaches}
Bayesian methods~\citep{gilks1996introducing,Goldstein1995} are often used for parameter estimation. Such models result in a distribution over parameters rather than point estimates, which can result in greater robustness to outliers and small sample sizes~\citep{miaou2003modeling}.
The empirical Bayes method (also known as maximum marginal likelihood) has been used in traffic engineering~\citep{Hauer1986,Hauer1992,Hauer1983,Heydecker2001} (the method as applied to crash prediction is explained particularly well by~\citeauthor{Hauer2002}~\citep{Hauer2002}).
Bayesian modelling techniques have also been used to assess potential risk factors of spatial regions~\citep{MacNab2004,pettet2017incident} and to estimate expected crash frequencies~\citep{aguero2009bayesian}.

Hierarchical Bayesian estimation of safety performance models have also been explored over the last two decades~\citep{Park2007,Ma2008,lord2008effects,miaou2005bayesian,Schluter1997,Davis2001}. 
Recently, the Poisson-gamma and Poisson-lognormal models have also been estimated using Bayesian methods~\citep{Quddus2008,akinCrashBinomial,basu2017regression,ladron2004forecasting,caliendo2007crash,ackaah2011crash,dissanayake2006statistical,Shirazi2019, Aguero-ValverdeJonathan2013FBPg, Khazraee2018}.
A caveat regarding Bayesian models is that the crucial choice of priors in the predictive models. The underlying information for designing priors might be available from previous models, engineering judgement, etc., and prior distributions can also be chosen to be non-informative or weakly informative. An important investigation in this context, specifically regarding crash prediction, has been done by \citeauthor{song2006bayesian}~\citep{song2006bayesian}, who study the performance of various Bayesian multivariate spatial models with different prior distributions.
It has also been shown that using non-informative priors may result in a high bias for the dispersion parameter in models, especially with small sample sizes~\citep{Park2010a}.

\subsubsection{Data Mining Approaches}
With improved sensor technology and easier storage, data-mining methods have successfully been used for crash prediction. 
Random forests~\citep{Abdel-Aty2008,Yu2014}, support vector machines~\citep{doi:10.3141/2024-11,Li2008,Yu2013}, and neural networks~\citep{Pande2006,Abdelwahab2002,Chang2005,Riviere2006} have recently been used to model crashes. Bayesian neural networks have also been explored, which address over-fitting of neural-networks in crash modeling~\citep{Xie2007}.
Deep learning techniques have also been used in various studies~\citep{zhu2018use,Bao2019}. One model that may be of interest to practitioners was developed by \citeauthor{basak2019analyzing}~\citep{basak2019analyzing}, who used a spatio-temporal convolution long short-term memory network (LSTM) to predict short-term crash risks, including propagation of traffic congestion. While the network structure was a combination of various complex networks, the accuracy of hourly predictions was limited, which highlights the inherent difficulty of predicting crash frequency at low temporal and spatial resolutions. It also makes a case against the use of complex models in this domain because they are harder to generalize.

\newtext{\subsubsection{Model Stacking}}  
\newtext{Ensemble methods use multiple trained models to improve prediction compared to what can be obtained from individual models. The most straightforward approach is averaging the prediction of two or more models. However, a better approach is to use a meta-learning algorithm to learn the best combination of the predictions from multiple models, which is known as stacking or stacked generalization \citep{alma991043603404303276}. Big data and the surge in availability of computational resources have paved for more sophisticated approaches such as model stacking in incident prediction. Various stacking models with different numbers of layers and assorted types of models \citep{iqbal2021efficient, tang2019crash, xiao2019svm, ma2021analytic,behura2020road, singh2018deep, chen2018sdcae} have been used during recent years to predict and detect incidents. The main caveats of using ensemble models are overfitting and the data size required for testing and training.} 

\newtext{\subsubsection{Modeling Time to Incidents}
A somewhat different approach for predicting emergency incidents is to directly model inter-incident time as a function of relevant covariates. In this case, the variable $X$ corresponds to the time between consecutive incidents. \citeauthor{mukhopadhyayAAMAS17}~\citep{mukhopadhyayAAMAS17} describe an example of such models by using uncensored (parametric) survival models to estimate time between accidents. It has been since used to model different incident types~\citep{pettet2017incident,MukhopadhyayICCPS,mukhopadhyayAAMAS18}. A key advantage of such methods is that planning problems are often modeled as continuous-time processes, and as a result, the incident prediction models can be easily used by planning models.
While time-based models are not the most commonly used approaches to model the occurrence of crashes, continuous-time models are often used for other purposes in ERM pipelines. For example, such models are widely used for predicting the duration of crashes and the delay that crashes cause in traffic and congestion~\citep{jiang2014traffic,tajtehranifard2016motorway,li2014traffic,chung2012methodological,basak2019data,6012554}. Hazard-based approaches have also been used to evaluate the time it takes to report, respond to, and clear incidents~\citep{nam2000exploratory}. While such algorithmic directions of work are crucial to the overall ERM pipeline, they lie outside the scope of this paper.}

\newtext{Another way to directly model time between incidents is to use time-series based forecasting. While approaches like survival models assume that inter-dependencies between successive incidents (or related in the feature space) can be modeled by designing appropriate features, time-series based approaches explicitly consider that consecutive observations are statistically \textit{dependent}. Typically, algorithmic approaches applicable to stationary time-series (defined as a series whose mean, variance, and auto-correlation are constant over time) have been used for forecasting roadway accidents~\citep{al1995time,khasnabis1989use,al2019diagnostic}. While non-stationary time-series data is more common in practice, such data can typically be converted to a stationary series by differencing~\citep{al1995time}.  The combination of time-series and data-mining based approaches have also been explored for forecasting traffic accidents~\citep{shao2019traffic}.}

\newtext{\subsubsection{Other considerations}}


\newtext{Most approaches to incident prediction assume that estimated model parameters do not change over time -- i.e. the parameters are temporally stable. However, several studies have found temporal instability in incident and injury-severity models' parameter estimates~\citep{behnood2015temporal, venkataraman2016transferability, marcoux2018evaluating, alnawmasi2019statistical, al2020temporal, islam2020temporal}. There are many reasons to expect model paramters to shift over time. Driver behavior has been shown to be influenced by factors such as macroeconomic conditions, cognitive biases that affect risk perception, and drivers’ attitudes toward safety~\citep{mannering2018temporal}. All of these factors change over time, suggesting that driver behavior is also temporally unstable. Additionally, the dynamics of urban environments evolve due to factors such as population shifts and roadway construction. It is important that such changes are taken into account by forecasting methodologies. Recently, the development of online models for predicting accidents has been explored that update learned models continuously using incoming streams of data~\citep{MukhopadhyayICCPS}. See \citeauthor{mannering2018temporal}'s work for a detailed discussion of the potential causes and implications of temporal instability in accident data~\citep{mannering2018temporal}.}

\newtext{An important consideration when evaluating various modeling approaches is the ability of each to reveal underlying causal relationships between features and the risk of incident occurrence. Often, there is a tradeoff between a model’s causal inference ability, scalability to large datasets, and predictive capability~\citep{mannering2020big}. To properly understand causality, statistical models must consider factors such as potential endogeneity in the data (discussed in section~\ref{sec:pred:features:other_considerations}) and unobserved heterogeneity with techniques such as random parameter models (section~\ref{sec:pred:models:rp}). Unfortunately, it is challenging to apply these methods to large datasets due to the complexity of estimating their parameters. Data-mining methods, on the other hand, scale very well to big-data applications, and have shown excellent predictive performance. However, this comes at the cost of causal inference, and their black-box nature makes it difficult to separate correlation from causation. Practitioners should be aware of these tradeoffs, and chose a modeling approach with strengths that align with the goals of their analysis. }


As incidents like accidents evolve in space and time, it is particularly important to identify the spatial and temporal resolutions that predictive models can accommodate. Naturally, changes in the degree of discretization affect the distributions of the dependent and the independent variables. On one hand, since forecasting the exact time and location of incidents like crashes is virtually impossible, high-resolution models are very difficult to construct. On the other hand, reducing the resolution may result in aggregation bias and unobserved heterogeneity~\citep{washington2020statistical}. A specific problem with fine-grained spatial and temporal discretization is the prevalence of zero counts in the resulting data, which might pose problems with convergence while statistical learning~\citep{lord2005poisson}. Similarly, it has also been shown that increasing the resolution of spatial and temporal discretization can lower the accuracy of various methods, including deep learning, tree-based, and econometric models~\citep{Bao2019}.
As a result, it is crucial that designers balance the spatial and temporal discretization of their models according to the specific needs of an area and explore the sensitivity of the model with changes in resolution.

\subsection{Feature Selection} \label{sec:FeatureSelection}



\newtext{ 
An important part of developing predictive models is data collection and feature engineering. Since various factors are involved in causing an accident, an important step in accident prediction pipeline is to collect as much as data as possible about relevant determinants. Collecting and cleaning data might be very challenging due to the size and incompatibility of datasets from different resources~\citep{vazirizade2021learning}. Furthermore, some micro-level features such as age of the drive or type of the car might not be available due to privacy and legal reasons.  In general, the performance of models strongly depends on the selected features. Consequently, they should be chosen strategically. Missing relevant explanatory features may result in an inaccurate model. On the other hand, including too many features requires more computational resources and may cause overfitting and erroneous prediction.}

\newtext{
Different approaches have been used to select a subset of all available features. Filter-based methods \citep{durduran2010decision}, wrapper methods \citep{lee2010computerized, tambouratzis2014maximising, ke2017missing, krishnaveni2011perspective}, embedded methods \citep{chen2019key, fu2019titan}, and combination of multiple methods \citep{wang2020crash, haruna2019discrete,shanthi2012feature, ramani2016learning} are the most common approaches for feature selection. SHAP (SHapley Additive exPlanations), a framework originally used for model interpretability, has also be used for evaluating the importance of the features in the model~\citep{parsa2020toward,wen2021quantifying}.}

In general, features for accident prediction can be categorized into temporal, spatial, or a combination of both. For example, one can choose to use time of day as a feature in order to understand how it affects accident rates. This is an example of a temporal feature. The geometry of a specific road segment, on the other hand, is a spatial feature, as it is a characteristic property of a particular spatial unit. Spatio-temporal features measure spatial properties that change with time. For example, traffic congestion in a specific part of the city falls under this category since it is characterized by both space and time. Unfortunately, not all of the underlying factors involved in an accident are measurable. The features available for crash analysis are usually restricted to the information on the crash report, weather and environmental conditions, roadway geometry, and traffic information. 
It is also possible to categorize features into static or dynamic~\citep{Qi2007}, but we choose to follow the categorization with respect to spatio-temporal characteristics of the features.

\subsubsection{Temporal Features}
Weather~\citep{songchitruksa2006assessing,mukhopadhyayAAMAS17,Qi2007} and visibility range~\citep{Abdel-Aty2012} have been proven to be useful in predicting accident rates, especially features like fog, rain, and snow. Weather data can also include seasonality features, temperature, light, etc. Time of the day and day of the week are also important predictors of accident rates~\citep{mukhopadhyayAAMAS17,Huang2008,Qi2007}.

\subsubsection{Spatial Features}
Roadway geometry is also known to be an effective predictor of crash frequency~\citep{poch1996negative,Khazraee2018,shankar1995effect,Chin2003}. The most commonly used features in this regard are the number of lanes, annual average daily traffic (AADT), segment length, width of the lanes, features regarding shoulders, horizontal turns and slopes \newtext{ ~\citep{Ma2008,Zeng2017,wen2021quantifying}}, the presence of uncontrolled left-turn lane, the presence of bus stops and surveillance cameras,  median widths, speed limit~\citep{Khazraee2018,Huang2008}, and features specific to intersections~\citep{Huang2008,Chin2003}. \newtext{Population density \citep{parsa2020toward}, road density \citep{Bao2019}, and socio-economic} features can also be important predictors of accidents rates, for example, the density of the bars in a region has been used in crash prediction, in particular hit and run accidents~\citep{Kuo2020}.

\subsubsection{Spatio-Temporal Features}
Crashes exhibit strong spatial-temporal incident correlation. Past incidents are an important predictor of future incidents. For example, areas that have typically experienced a relatively high concentration of incidents in the recent past are more likely to have incidents in the future~\citep{mukhopadhyayAAMAS17,mukhopadhyayAAMAS18,MukhopadhyayICCPS}. The average speed of vehicles naturally serves as a predictor for the likelihood of crashes~\citep{Shi2015}. The role of traffic congestion has been studied in the context of crash analysis. Traffic congestion has been shown to increase the likelihood of rear-end crash~\citep{Shi2015}, while there have been studies showing that congestion has no or negative effect on crash frequency~\citep{Wang2009,baruya1998speed}. Several other features like peak hour~\citep{Qi2007} and traffic volume~\citep{Xie2007}, which are indirect measures of congestion, can also be used as covariates to model crash occurrence.

\subsubsection{Other Considerations} \label{sec:pred:features:other_considerations}
In addition to accounting for different types of features (spatial, temporal, spatio-temporal), there are various considerations that need to be addressed. One of the most important but often ignored challenges is endogeneity. The source of endogeneity can be broadly classified as omitted variables (unobserved heterogeneity), simultaneity, and biased sampling. Unobserved heterogeneity refers to features that are not recorded in the data but that affect the occurrence of accidents or are correlated to other observed features~\citep{mannering2014analytic}. For instance, a recent study found that the positive correlation effect between tangent segments and a wider left shoulder signifies increased crash occurrence on multilane highways. This can be the result of heterogeneous driving behavior where the risk perception of driving on tangent segments with wider left shoulders can lead drivers to speed~\citep{saeed2019analyzing}. Simultaneity arises when the explanatory variable causes the response variable and the response variable causes the explanatory variable. The classic example of simultaneity is the influence of ice-warning signs on increasing accident rates. In reality, these signs are installed in locations where the accident rate is high due to icy surfaces~\citep{Lord2010}. Finally, the data collected for accident prediction might be biased due to various reasons. Under reporting of crashes with no severe injuries is a prime example of biased sampling~\citep{yasmin2014latent}. Failure to consider the aforementioned challenges can lead to biased parameter estimates and incorrect inferences on accident rates.

\subsection{Incident Severity}\label{sec:AccidentSeverity}



Severity of accidents plays a crucial role in planning approaches for allocation as well as for dispatching resources when incidents occur. Naturally, decision-makers plan to prioritize incidents with higher severity over the ones with relatively lower severity. Since it is difficult to gauge the severity of an incident based on a call for assistance, it is common in practice to dispatch the responder closest to the scene of the incident. However, understanding spatial and temporal patterns in severity and its relationship with incident occurrence models is crucial in optimizing the allocation of responders. Understanding covariates that affect severity, and creating models for predicting severity of crashes have attracted a lot of attention. While there are different definitions of severity, it can usually be categorized into five levels: 1) no-injury or just property damage, 2) possible injury, 3) non-incapacitating injury, 4) incapacitating injury, and 5) fatal~\citep{Savolainen2011}. Most of the prior work in severity prediction has focused on using a similar ordinal categorization. \citeauthor{Savolainen2011}~\citep{Savolainen2011} present a detailed review regarding modeling severity of accidents, which is self-contained, complete, and comprehensive. Much of this section is informed by their work; we identify crucial insights from it and also focus on models that have been introduced since then. 

Let incident severity be represented by the random variable $K$. From the perspective of the formulation in section \ref{sec:formulation}, designing models for incident severity can be represented in two ways. First, there is significant work on creating marginal models over severity. These models have the form $h(K \mid w,\theta)$, where $h$ is a distribution over $K$, $w$ is a set of covariates that impact incident severity, and $\theta$ denotes the model parameters. Note that $w$ could include information about the crash itself, such as information from post-crash reports. The other approach is to model a joint distribution that governs incident occurrence and the resulting severity. In this scenario, given incident data, the decision-maker seeks to learn a joint distribution over incident occurrence and severity, which can be represented by $h(X,K \mid w,\theta)$.

The relationship between traffic flow and accident severity is well-explored \citep{jadaan1992relationships,martin2002relationship,turner1986motorway,mcguigan1987examination,hall1990rural}. Crash severity has been explored using multinomial logit and probit models~\citep{Ye2014,khattak1998role,shibata1994risk,fan2016identifying,Kockelman2002}, decision trees \citep{chang2006analysis}, random forests \citep{Zheng2018,iranitalab2017comparison,mafi2018machine}, and neural networks \citep{rezaie2011prediction,chimba2009prediction}. Additionally, studies have used the RP models (correlated and uncorrelated) and Bayesian model to evaluate the impact of roadways design and weather on crash injury-severity~\citep{fountas2018analysis,Huang2008}.

One natural way to account for correlation between crash frequency and severity is to learn an independent regression model for each category of severity. Multiple regression models~\citep{bijleveld2005covariance,ma2006bayesian,song2006bayesian} as well as neural networks~\citep{zeng2016modeling} have been used to this end.
Although such a paradigm captures inherent correlation (to some extent) between incident arrival and severity, it does not model an explicit joint distribution. \citeauthor{mukhopadhyayAAMAS17}~\citep{mukhopadhyayAAMAS17} present an approach that forms a bridge between marginal and joint models. They assume that the joint distribution can be decomposed into a marginal distribution over incident arrival, followed by a conditional distribution over severity given incident arrival. 

In the last two decades, there has also been significant interest in jointly modeling incident arrival (frequency) and severity~\citep{ma2006bayesian,Ma2008,Park2007,aguero2009bayesian,pei2011joint}. This includes multivariate Poisson regression~\citep{ma2006bayesian} and multivariate Poisson log-normal regression models~\citep{Park2007}. \citeauthor{pei2011joint}~\citep{pei2011joint} model the joint distribution explicitly and use a fully Bayesian approach to learn the model. While such models are promising, a crucial (potential) limitation is identified by~\citeauthor{Savolainen2011}~\citep{Savolainen2011}. Jointly modeling crash arrival and severity limits the use of data related to the specific crash while learning the model. On the other hand, marginal models can use detailed post-crash data to infer insights about severity~\citep{Savolainen2011}.  It is worth mentioning that factors such as the type of the vehicle, age of the driver, and using seat belt are useful in severity prediction~\citep{8842628}, which are usually not included in the collected environmental data.

Finally, there are two orthogonal areas of work in severity prediction that can be combined with both marginal models and joint models. The first approach is rather recent and focuses to identify spatial relationships between different levels of severity~\citep{Kuo2020}. The other approach seeks to tackle inherent heterogeneity in crash data by identifying clusters of incidents (not necessarily spatial) to better understand the relationship between crash data and covariates~\citep{sun2019pedestrian,sasidharan2015exploring,sivasankaran2020exploring}.

\subsection{Key Takeaways}

\begin{figure*}[t]
\begin{center}
\vspace{-0.1in}
\includegraphics[height=5cm]{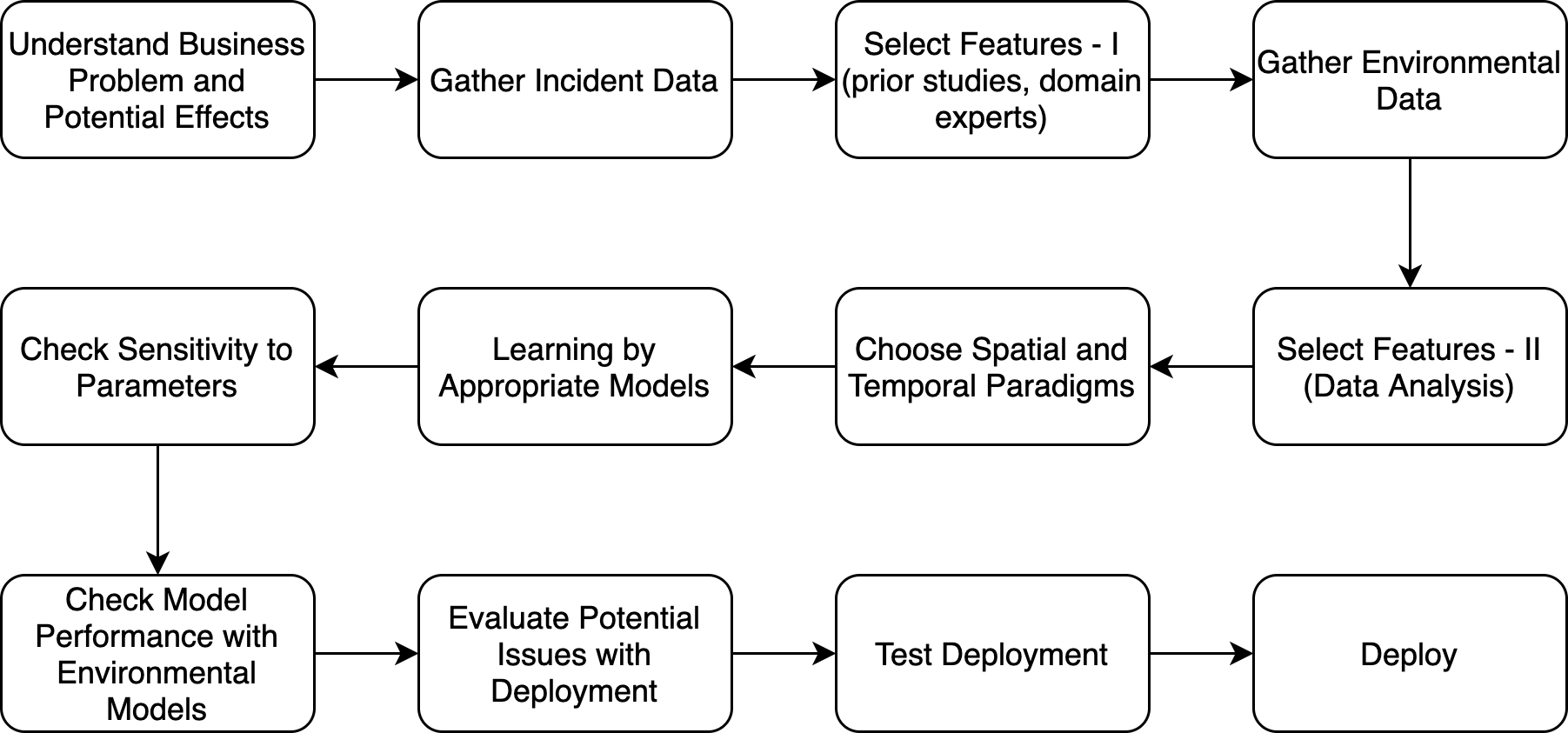}
\caption{Incident Prediction Model Design Pipeline}
\label{fig:predSteps}
\end{center}
\vspace{-0.3in}
\end{figure*}

%

Based on the approaches surveyed, we summarize steps that practitioners and model designers should take in Figure~\ref{fig:predSteps}. Specifically, we recommend practitioners, model designers, and planners to:
\begin{enumerate}[noitemsep]
    \item Be aware of advances made in predictive modeling in the context of different types of incidents.
    \item Seek the help of domain experts (researchers, fire-fighters, policy-makers, etc.) to design the feature space $w$, which is a crucial factor in the performance of predictive models.
    \item Start by using well-defined paradigms that have been shown to work on multiple datasets (e.g. hierarchical Poisson models and zero-inflated models).
    \item Be aware of flaws and shortcomings of models, and carefully evaluate the possible costs of inaccurate predictive models.
    \item Be aware that many different kinds of predictive models have been used to model crash occurrence, and models can be sensitive to a variety of factors like the granularity of spatial discretization, choice of covariates, etc. As a result, it is important that practitioners test multiple models and evaluate their comparative performance to understand the model that best suits a particular scenario. 
\end{enumerate}

\section{Event Extraction}\label{sec:extraction}

Response to incidents like accidents and medical emergencies must be dispatched as soon as possible. For decades, the pipeline depended on a human reporting the incident, after which responders were dispatched to the scene. However, with the advent of a variety of sensors available in smart cities and the wide corpus of information on social media, it is now possible to detect incidents before they are reported. For example, consider a fire in an urban area. An observer might share the observation on social media, or the incident might be captured by video cameras installed at traffic intersections. The goal of event extraction is to use such data to detect the occurrence of incidents in order to reduce the overall time for response.

Event extraction algorithms seek to identify as much information about an incident as possible, with a focus on specific details such as the location, time, and the agents involved. For different events, the \textit{who, where, what, when}, and \textit{why} information vary a lot, but similar events usually share the same event template. For example, in an \textit{accident} event, there are usually \textit{entities} involved in the accident, the \textit{location} of the event, and its \textit{time} of occurrence. 
Researchers in Natural Language Processing (NLP) community have developed event ontologies to define the templates for various events~\citep{baker1998berkeley,schuler2005verbnet,doddington2004automatic}. Event extraction can then be defined as a task to convert unstructured data into event-centered structured data based on a specific event ontology. This is greatly beneficial for first responders since it is much easier for users to manage and query structured data. We focus our attention on two major categories of models for event extraction (EE): (1) EE from textual information, and (2) EE from multimedia information. \newtext{We point out that there has also been recent work in extracting information about events by using crowdsourcing applications such as Waze~\citep{yasas2021}, who focus on optimizing the balance of practitioner-centric parameters (e.g., spatial and temporal localization of alerts) with learning outcomes (e.g., accuracy).}

\subsection{Event Extraction from Textual Information}
The goal of EE using NLP is
to identify events from text and classify them into pre-defined categories, as well as identify event participants (for example, the victim in an accident), and event attributes (for example, the location and time of an event).

\begin{figure*}[ht]
    \centering
    \includegraphics[width=\textwidth]{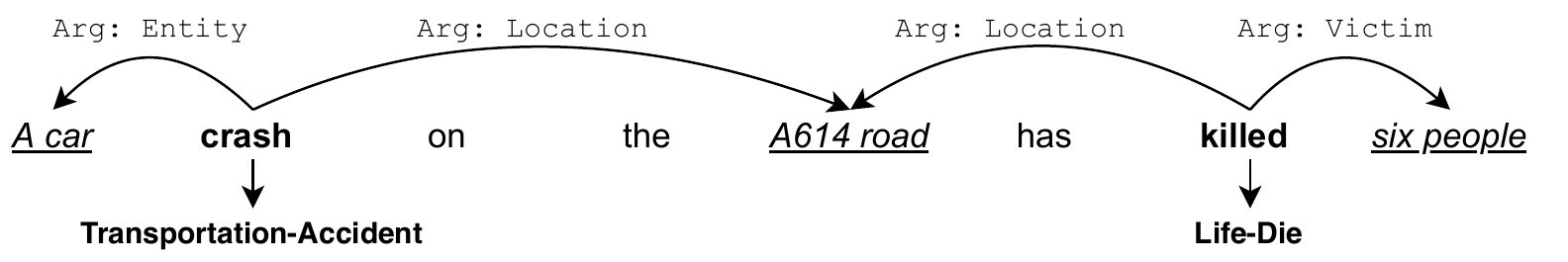}
    \caption{An example of Transportation-Accident and Life-Die events and their arguments.}
    \label{fig:ee}
\end{figure*}

There are two subtasks in  EE: (1) trigger classification, which aims to identify the word/phrase that clearly expresses the occurrence of an event and seeks to classify it based on pre-defined categories. For example, the word `crash' in Figure~\ref{fig:ee} is the trigger word of a Transportation-Accident event; (2) event argument classification, which aims to identify the event participants and event attributes. For example, `A614 road', and `a car' are Location argument and Entity argument of the Transportation-Accident event respectively (Figure~\ref{fig:ee}).

The early work on EE mainly relies on feature engineering and adapts a pipeline framework~\citep{grishman2005nyu,ahn2006stages,ji2008refining,hong2011using,mcclosky2011event}. The input sentence is first tokenized into a sequence of tokens. For each token, various features are used, which can be divided into three categories: (1) lexical features such as n-grams, lemma, and synonyms of tokens, and brown clusters, (2) syntactic features such as dependent and governor tokens, and (3) entity features such as entity type. Then a classifier is trained based on statistical models (for example, logistic regression) to predict event triggers. Subsequently,  another classifier is trained to predict the event arguments. During inference, the argument classifier receives the prediction from the trigger classifier as input; therefore, the errors from the former classifier are easily propagated into the latter one.
To mitigate the error propagation issue in pipeline models, a joint model can be learned. For example, \citeauthor{li2013joint}~\citep{li2013joint} proposed using structured perceptron~\citep{collins2002discriminative} with beam search to learn a joint model by leveraging the dependencies between arguments and triggers.

The limitation of models based on feature engineering is that they rely on handcrafted features and language-specific resources such as part-of-speech (POS) taggers and dependency parsers. As a result, it is hard to adapt such models to new languages or domains. This problem manifests frequently in working with textual data from social media because of language variations and informal grammar used in such platforms. Indeed, POS taggers and dependency parsers perform much worse in the context of social media than in the structured news domain. With the idea of using word embeddings~\citep{mikolov2013distributed}, deep neural networks have become an attractive choice for researchers because no handcrafted features are required. For example, \citeauthor{chen2015event}~\citep{chen2015event} applied convolutional neural networks (CNN) in a two-step pipeline system, in which the tokens are converted into pre-trained word vectors. 
\newtext{\citeauthor{nguyen2016joint}~\citep{nguyen2016joint} proposed a joint framework with bidirectional recurrent neural networks.}

However, a potential issue with the use of simple CNN models is that they can hardly capture the syntactic relations between words, which are very important features for argument classification. \citeauthor{nguyen2018graph}~\citep{nguyen2018graph} applied Graph Convolutional Networks (GNN) based on dependency trees to generate word representations by leveraging the information from other words with close syntactic relations.

\newtext{The models mentioned above first identify event trigger and entity and then perform argument role classification. There has also been recent work that bypasses the entity recognition step. For example, \citeauthor{wadden2019entity}~\citep{wadden2019entity} learn entity span representations instead of explicitly labeling the entity using BIO schema, and \citeauthor{du2020event}~\citep{du2020event} model event extraction as question answering and extracts the spans of event arguments with certain role types.}

\subsection{Event Extraction from Multimedia Information}
Event extraction is also possible from video and image data~\citep{gan2015devnet,chang2016bi,ma2017joint,li2020cross}. In the context of emergency response, cities are increasingly trying to use traffic cameras to track and monitor congestion and detect incidents that need response. \citeauthor{gan2015devnet}~\citep{gan2015devnet} applied a CNN pre-trained on Imagenet~\citep{deng2009imagenet} to perform keyframe detection. A weighted-sum of the representations learned from different multimedia archives to generate better representations for event videos has also been used~\citep{chang2016bi}. Lower level video attributes learned from external multimedia datasets can also be used to improve complex event detection in videos~\citep{ma2017joint}. A potential drawback of such approaches is that they fail to be applicable in complex scenarios that have event arguments.
\newtext{More recently, \citeauthor{li2020cross}~\citep{li2020cross} propose a new task for multimedia EE, in which they extend the EE task from textual setting into the multimedia setting and target complex events. The extension is done by applying weakly supervised training to project the structured semantic representations from textual and visual data into a common space.} Event extraction is an emerging field of research and with the rapid growth of smart and connected communities, it promises to play a vital role in the design of principled ERM pipelines.

\section{Responder Allocation and Dispatch}\label{sec:ResponderAllocationandDispatch}

There are two important steps in an ERM system that come into effect \textit{after} the decision-maker gains an understanding of when and where incidents are likely to happen. These involve allocating resources (also referred to as the stationing problem \citep{pettet2020algorithmic}) in anticipation of incidents and dispatching resources when calls for service are received (also referred to as the stationing problem). While prediction problems are primarily formulated as \textit{learning} problems, allocation, and response are commonly modeled as optimization problems. As discussed in section \ref{sec:formulation}, an allocation or dispatch problem can be represented as $\max_{y} G(y \mid f)$, where $y$ represents the decision variable, $G$ is a reward function chosen by the decision-maker, and $f$ is the model of incident occurrence. For allocation problems, $y$ typically refers to the location of emergency responders in space. For dispatch problems, the decision variable is a mapping between responders and specific calls for service. 

The distinction between allocation and dispatch problems can be hazy since any solution to the allocation problem implicitly creates a policy for response. For example, consider an algorithm that allocates ambulances across an urban area in a manner that minimizes expected response times to incidents according to an incident arrival model $f$. Now, when an incident occurs in the jurisdiction of a specific station, naturally, a responder (if available) is dispatched from the station, without the need for an explicit dispatch model. While this is generally true for allocation models, there are finer subtleties involved. As noted by Mukhopadyay et.al. ~\citep{mukhopadhyayAAMAS17,mukhopadhyayAAMAS18,MukhopadhyayICCPS}, implicit response strategies are not always optimal. For example, consider a situation where an incident occurs close to a station that has no available responders. Should the incident enter a waiting queue? How does the potential severity of the concerned incident affect this decision? If a nearby station has a free responder, should it be dispatched? How do response time guarantees from the allocation model change in such scenarios? Answering such questions is critical for an efficient ERM system, which motivates the design of principled approaches for dispatching responders. This section discusses prior work on the problems of responder allocation and response.

We first introduce the metrics used to allocate emergency response stations and responders. The three most common metrics are coverage~\citep{toregas_location_1971,church1974maximal, gendreau_solving_1997}, distance between facilities and demand locations~\citep{MukhopadhyayICCPS}, and patient survival~\citep{erkut_ambulance_2008, knight_ambulance_2012, mccormack_simulation_2015}. \textit{Coverage} measures the proportion of spatial locations that are within some predefined distance of the responders (or depots). It is measured with respect to demand nodes, which are discretized spatial units that can potentially generate calls for service. Of the three metrics, it is the most straightforward to examine as it is generally binary. A demand node is considered to be \textit{covered} by some facility if it is within the predefined distance, and otherwise considered to be uncovered. It also lines up well with the broader objective of many EMS providers, which is to limit the number of calls that are responded to \textit{late}, i.e. that have a response time higher than some threshold (the distance often serves as a proxy for the response time, for example see \citeauthor{mukhopadhyayAAMAS17}~\citep{mukhopadhyayAAMAS17}). These factors contributed to coverage being a prevalent metric in early EMS allocation research.

The distance between potential demand nodes and their nearest facilities is another metric that can be used for optimization of the spatial distribution of stations and responders. These metrics are more difficult to use since they are not binary, but recent advances in computational capabilities have made them more accessible. Both coverage and distance to potential demand locations actually approximate the true objective of EMS policies, which is increasing patient survival. \citeauthor{erkut_ambulance_2008}~\citep{erkut_ambulance_2008} argue that it is more appropriate to use expected patient survival directly by incorporating a survival function that captures the relationship between response times and survival rates. 

Most early ERM allocation approaches modeled the allocation problem as an integer or linear optimization problem \citep{toregas_location_1971, church1974maximal, gendreau_solving_1997}. These models are relatively straightforward and can be solved by a large body of optimization techniques. Exact methods such as branch-and-bound have been applied to small instances of the problem \citep{swoveland1973simulation, marianov_queuing_1994} but do not easily scale to realistic environments. As a result, most prior work relies on heuristic approaches, such as genetic algorithms \citep{jia_modeling_2007, rajagopalan_developing_2007} and tabu search \citep{gendreau_solving_1997, rajagopalan_developing_2007, gendreau_dynamic_2001, rajagopalan_multiperiod_2008}. Recently, decision theoretic models such as Markov decision processes (MDPs) have gained traction as efficient solution methods have evolved \citep{maxwell_ambulance_2009, MukhopadhyayICCPS}. 

\subsection{Coverage Models}

\newtext{
\begin{table*}[ht]
\centering
\resizebox{\textwidth}{!}{%
\begin{tabular}{llll}
\hline
\multicolumn{1}{|l|}{Approach}                                                                       & \multicolumn{1}{l|}{Objective}                                                                                                                             & \multicolumn{1}{l|}{Description}                                                                                                                                                                                                                                                          & \multicolumn{1}{l|}{Reference}                     \\ \hline
\multicolumn{1}{|l|}{LSCP}                                                                           & \multicolumn{1}{l|}{\begin{tabular}[c]{@{}l@{}}Minimize number of facilities \\ to completely cover demand.\end{tabular}}                                  & \multicolumn{1}{l|}{\begin{tabular}[c]{@{}l@{}}Finds a lower bound on the number of facilities\\ needed for a given coverage standard. \\ Assumes facilities can service all demand they cover.\end{tabular}}                                                                       & \multicolumn{1}{l|}{\cite{toregas_location_1971}}  \\ \hline
\multicolumn{1}{|l|}{\begin{tabular}[c]{@{}l@{}}LSCP with \\ continuous \\ regions\end{tabular}}     & \multicolumn{1}{l|}{\begin{tabular}[c]{@{}l@{}}Minimize number of facilities \\ to completely cover demand.\end{tabular}}                                  & \multicolumn{1}{l|}{\begin{tabular}[c]{@{}l@{}}Extends LSCP to consider spatially continuous \\ regions rather than discrete demand points.\end{tabular}}                                                                                                                           & \multicolumn{1}{l|}{\cite{aly_probabilistic_1978}} \\ \hline
\multicolumn{1}{|l|}{MCLP}                                                                           & \multicolumn{1}{l|}{\begin{tabular}[c]{@{}l@{}}Maximize demand covered \\ by given number of facilities.\end{tabular}}                                     & \multicolumn{1}{l|}{\begin{tabular}[c]{@{}l@{}}Represents problems where the number of facilities \\ is heavily constrained by costs. \\ Assumes facilities can service all demand they cover.\end{tabular}}                                                                        & \multicolumn{1}{l|}{\cite{church1974maximal}}      \\ \hline
\multicolumn{1}{|l|}{\begin{tabular}[c]{@{}l@{}}MCLP with \\ multiple\\ quality levels\end{tabular}} & \multicolumn{1}{l|}{\begin{tabular}[c]{@{}l@{}}Maximize the quality-weighted \\ demand covered by facilities with \\ various quality levels.\end{tabular}} & \multicolumn{1}{l|}{\begin{tabular}[c]{@{}l@{}}Extends MCLP by introducing multiple quality levels\\ representing each facility's available services or distance\\ to demand points.\end{tabular}}                                                                                  & \multicolumn{1}{l|}{\cite{jia_modeling_2007}}      \\ \hline
\multicolumn{1}{|l|}{MSLP}                                                                           & \multicolumn{1}{l|}{\begin{tabular}[c]{@{}l@{}}Maximize patient survival with \\ given number of facilities.\end{tabular}}                                 & \multicolumn{1}{l|}{\begin{tabular}[c]{@{}l@{}}Extends MCLP by mapping response times \\ to patient survival rates.\end{tabular}}                                                                                                                                                   & \multicolumn{1}{l|}{\cite{erkut_ambulance_2008}}   \\ \hline
\multicolumn{1}{|l|}{DSM}                                                                            & \multicolumn{1}{l|}{\begin{tabular}[c]{@{}l@{}}Maximize the demand covered at least \\ twice with available facilities.\end{tabular}}                      & \multicolumn{1}{l|}{\begin{tabular}[c]{@{}l@{}}Accounts for facilities being busy by ensuring \\ demand is covered by at least two facilities.\end{tabular}}                                                                                                                        & \multicolumn{1}{l|}{\cite{gendreau_solving_1997}}  \\ \hline
\multicolumn{1}{|l|}{MEXCLP}                                                                         & \multicolumn{1}{l|}{\begin{tabular}[c]{@{}l@{}}Maximize demand covered\\ by given number of facilities.\end{tabular}}                                      & \multicolumn{1}{l|}{\begin{tabular}[c]{@{}l@{}}Extends MCLP to account for station availability.\\ Assumes all facilities have same busy probabilities and\\ that facilities act independently.\end{tabular}}                                                                       & \multicolumn{1}{l|}{\cite{daskin_maximum_1983}}    \\ \hline
\multicolumn{1}{|l|}{TIMEXCLP}                                                                       & \multicolumn{1}{l|}{\begin{tabular}[c]{@{}l@{}}Maximize demand covered\\ by given number of facilities.\end{tabular}}                                      & \multicolumn{1}{l|}{\begin{tabular}[c]{@{}l@{}}Extends MEXCLP by introducing temporal variations\\ in travel times between locations.\end{tabular}}                                                                                                                                 & \multicolumn{1}{l|}{\cite{repede_developing_1994}} \\ \hline
\multicolumn{1}{|l|}{AMEXCLP}                                                                        & \multicolumn{1}{l|}{\begin{tabular}[c]{@{}l@{}}Maximize demand covered\\ by given number of facilities.\end{tabular}}                                      & \multicolumn{1}{l|}{\begin{tabular}[c]{@{}l@{}}Extends MEXCLP, relaxing the assumption that \\ facilities are independent.\end{tabular}}                                                                                                                                            & \multicolumn{1}{l|}{\cite{batta_maximal_1989}}     \\ \hline
\multicolumn{1}{|l|}{MALP-I}                                                                         & \multicolumn{1}{l|}{\begin{tabular}[c]{@{}l@{}}Maximize demand that is covered\\ by facilities with a exogenously \\ specified probability.\end{tabular}}  & \multicolumn{1}{l|}{\begin{tabular}[c]{@{}l@{}}Assumes all facilities have same busy probabilities\\ and that facilities act independently.\end{tabular}}                                                                                                                           & \multicolumn{1}{l|}{\cite{revelle_maximum_1989-1}} \\ \hline
\multicolumn{1}{|l|}{MALP-II}                                                                        & \multicolumn{1}{l|}{\begin{tabular}[c]{@{}l@{}}Maximize demand that is covered\\ by facilities with a exogenously \\ specified probability.\end{tabular}}  & \multicolumn{1}{l|}{\begin{tabular}[c]{@{}l@{}}Relaxes assumption that all facilities have the same busy\\ probabilities. Busy probabilities are computed as a ratio\\ between demand generated at demand points and the \\ availability of facilities covering them.\end{tabular}} & \multicolumn{1}{l|}{\cite{revelle_maximum_1989-1}} \\ \hline
\multicolumn{1}{|l|}{QPLSCP}                                                                         & \multicolumn{1}{l|}{\begin{tabular}[c]{@{}l@{}}Maximize demand that is covered\\ by facilities with a exogenously \\ specified probability.\end{tabular}}  & \multicolumn{1}{l|}{\begin{tabular}[c]{@{}l@{}}Extends MALP-II by relaxing the assumption that \\ facilities are independent.\end{tabular}}                                                                                                                                         & \multicolumn{1}{l|}{\cite{marianov_queuing_1994}}  \\ \hline
                                                                                                     &                                                                                                                                                            &                                                                                                                                                                                                                                                                                     &                                                   
\end{tabular}%
}
\caption{\newtext{Coverage Models}}
\label{tab:coverage_models}
\end{table*}
}

Early allocation approaches also generally tackled \textit{static} allocation. Depots (also referred to as `facilities' or `stations') are assumed to be immobile, so the model determines the optimal locations for the depots without allowing for temporal redistribution. In such models, responders are often used synonymously with facilities. The two seminal static facility allocation models are the Location Set Covering Problem (LSCP) \citep{toregas_location_1971} and the Maximal Covering Location Problem (MCLP) \citep{church1974maximal}. Both models have similar assumptions, including that stations act independently, response is deterministic, that at most one ambulance is at each facility, and that there is one type of ambulance. The primary difference between the two is in the optimization objective. LSCP finds the least number of facilities that \textit{cover} all demand nodes, while MCLP maximizes the demand covered by a given number of facilities. LSCP can be useful for planning a lower bound on the number of facilities needed for a given coverage standard, while MCLP better captures the constraints of real world use cases where the number of facilities is heavily constrained by cost. It is also common to introduce constraints on secondary objectives like waiting times in optimization problems that seek to maximize coverage. For example, \citeauthor{silva2008locating}~\citep{silva2008locating} and \citeauthor{mukhopadhyayAAMAS17}~\citep{mukhopadhyayAAMAS17} define optimization frameworks for maximizing coverage with upper bounds on waiting times, and can accommodate different levels of incident severity.

There are a number of extensions to LSCP and MCLP, many of which relax some of their strong assumptions. \citeauthor{aly_probabilistic_1978}~\citep{aly_probabilistic_1978} consider a spatially continuous demand model, rather than the discrete demand nodes. \citeauthor{jia_modeling_2007}~\citep{jia_modeling_2007} introduce different quality levels for facilities (which can represent each facility's available services or equipment), with demand points having different coverage constraints for each level. \citeauthor{erkut_ambulance_2008}~\citep{erkut_ambulance_2008} incorporated a survival function into the optimization function of MCLP, which maps response times to survival rates, to formulate the Maximum Survival Location Problem (MSLP).

LSCP, MCLP, and many of their extensions all have a common shortcoming in that they assume deterministic system behavior in regards to response. Resources at a facility are considered to be always available, and the models assume that a station is able to service all demand nodes that it covers. In the real world, there are finite resources at each station, and calls from a specific demand node might need to be answered by a station other than the closest one. For example, it is common for other stations to respond to a call if the closest one is busy. One way to address this is by increasing the number of stations that cover each demand point, i.e. using a \textit{multiple coverage} metric. 

A key example is the Double Standard Model (DSM)~\citep{gendreau_solving_1997}, which incorporates two distance standards $r_1$ and $r_2$, where $r_1 < r_2$. The model adds the constraint that all demand must be covered within $r_2$, similarly to LSCP, ensuring that each point has \textit{some} coverage. It also specifies that some proportion $\alpha$ of the demand is covered within $r_1$. Given those constraints, the objective is to maximize the demand covered \textit{by at least two stations} within $r_1$. Essentially, this maximizes the demand nodes that have nearby facilities while ensuring that all demand nodes have adequate coverage. 
While this approach helps mitigate the issue of station unavailability, there can still be situations where both facilities covering some demand points are busy. Accounting for such situations requires modeling facility availability explicitly.  

There is a large body of research on probabilistic coverage models, which model the stochastic nature of station availability. Two foundational probabilistic models are the Maximum Expected Covering Location Model (MEXCLP) and Maximum Availability Location Problem (MALP). MEXCLP was introduced by \citeauthor{daskin_maximum_1983}~\citep{daskin_maximum_1983} and extends MCLP, modifying the optimization function to account for station availability. It assumes that each facility has the same probability of being busy, which simplifies computation but does not accurately represent the real world where facilities near incident hot spots are unavailable for a greater proportion of the time. Also, it inherits many of the assumptions of MCLP, and assumes that facilities act independently. MALP, proposed by \citeauthor{revelle_maximum_1989}~\citep{revelle_maximum_1989-1}, maximizes the demand covered by facilities with some exogenously specified probability. The first version, MALP-I~\citep{revelle_maximum_1989-1} is similar to MEXCLP in that it assumes equal probabilities for being busy for facilities. MALP-II~\citep{revelle_maximum_1989-1}, however, removes this assumption. The proportion of time that facilities are busy is computed as a ratio between the total demand generated by demand points and the availability of facilities covering them.

There have been several extensions to the above probabilistic models to relax some of their simplifying assumptions and make them better match the real world. TIMEXCLP, developed by \citeauthor{repede_developing_1994}~\citep{repede_developing_1994}, introduces temporal variations in travel times between points to MEXCLP. Adjusted MEXCLP (AMEXCLP) \citep{batta_maximal_1989} relaxes MEXCLP's assumption that facilities are independent by treating them as servers in a hypercube queuing system~\citep{larson_hypercube_1974} with equal busy factions. The Queuing Probabilistic Location Set Covering Problem (QPLSCP)~\citep{marianov_queuing_1994} makes a similar extension to MALP by computing each individual facility's busy fraction using a queuing model and feeding them into MALP-II. We summarize the different coverage models from prior work, their objective functions, and features in Table~\ref{tab:coverage_models}.

\subsection{Decision-Theoretic Models}
\newtext{Having discussed how the ERM planning problem can be framed as an explicit optimization problem, we now discuss an alternate approach which models the planning problem as a stochastic control problem, and then optimizes over the set of control choices to maximize expected reward.} The most commonly used model in this regard is the Markov decision process (MDP). MDP-s can be used as a general framework for sequential decision problems under uncertainty given a model of the concerned system~\citep{kochenderfer2015decision}. In such a formulation, an agent chooses an action at a given state of the system and receives a specific reward based on a pre-defined utility function. The system then transitions to a new state probabilistically. The \textit{Markovian} assumption means that the subsequent state depends only on the current state and the action taken. MDP-s have been used extensively to model the EMS dispatch process~\citep{carter1972response,keneally2016markov,mukhopadhyayAAMAS18,MukhopadhyayICCPS,mukhopadhyayAAMAS17,pettet2020algorithmic}.  

\citeauthor{carter1972response}~\citep{carter1972response} demonstrate one of the earliest examples of using an continuous-time MDP to aid emergency response by using a queuing model for calls. The general framework of such an approach has been used in several studies since then~\citep{mukhopadhyayAAMAS18,MukhopadhyayICCPS,mukhopadhyayAAMAS17}. \citeauthor{keneally2016markov}~\citep{keneally2016markov} also model the optimal dispatch problem as a continuous-time MDP and consider different levels of priorities for the incidents while dispatching. They assume that the state transition function for the EMS system can be expressed in closed-form and use canonical policy iteration to solve the problem. A shortcoming of such a model is that it assumes that state transitions follow a memoryless distribution. Real world transitions are not necessarily memoryless, and this is addressed by Mukhopadhyay  et.al.~\citep{mukhopadhyayAAMAS18}, who formulate the problem as a semi-Markovian decision problem (SMDP) instead. In the absence of closed-form expressions for state transitions, a black-box simulator can be used to learn an optimal dispatch policy. However, such an approach does not scale well to realistic scenarios. An approach to alleviate this problem is to focus on finding an action for the current state of the world instead of aiming to find a policy for the entire state-space~\citep{MukhopadhyayICCPS}. Given a generative model for the EMS system, heuristic search approaches like Monte-Carlo tree search (MCTS) can be used to find promising actions for the current state of the MDP~\citep{MukhopadhyayICCPS,pettet2020algorithmic}. An advantage of using MCTS is that the Markovian assumption can be relaxed, and high-fidelity simulators can be used to estimate utilities from different actions.

An important aspect of using decision-theoretic models is to carefully design the utility function. As discussed earlier, a threshold on response time can be used to penalize the number of calls that are responded to late. However, an explicit relationship between patient survival and response time can be directly used to design the reward function for MDP-s~\citep{bandara2012optimal}. It has also been highlighted that a consideration of priorities of calls is crucial to take into account while designing the utility function for decision-theoretic approaches to EMS~\citep{keneally2016markov,bandara2012optimal}.

\subsection{Dynamic and Proactive Reallocation}
A potential shortcoming of algorithmic dispatch approaches is important to ponder over. \newtext{Based on conversations with first responders, \citeauthor{pettet2020algorithmic}~\citep{pettet2020algorithmic} point out that the moral constraints in emergency response dictate that the nearest responder be dispatched to the scene of an incident, particularly when the severity of an incident cannot be gauged from the call for service. As a result, the decision variable that can be optimized exogenously is the location of responders in anticipation of incidents; on the other hand, choosing which responder to dispatch is done in a greedy manner.}
\citeauthor{pettet2020algorithmic}~\citep{pettet2020algorithmic} create an approach to optimize over the spatial distribution of responders \textit{between} incidents, while always dispatching the closest available responder to attend to incidents. This process alleviates two major issues. First, it does not waste crucial time \textit{after} an incident has occurred to optimize over  which responder to dispatch. Second, the moral constraint of always sending the closest responder to an incident is not violated. Dynamically reallocating emergency responders has actually been investigated earlier, to maximize the overall efficacy of response; such approaches are not necessarily motivated by the use of greedy dispatch policies. An approach to tackle the allocation problem dynamically is to formulate an integer-program and solve it in real-time every time reallocation needs to be made~\citep{maxwell2010approximate,kolesar1974algorithm}. Dynamic reallocation has also been addressed by decision-theoretic formulations, which seek to find optimal policies that govern when and where specific units should be moved. The utility function typically consists of expected time taken to serve requests in the long run. Optimal policies can be found by dynamic programming or approximate dynamic programming techniques~\citep{maxwell_approximate_2010, berman1981dynamic}.

\begin{figure*}[ht]
\centering
\begin{center}
\vspace{-0.1in}
\includegraphics[width=0.9\textwidth]{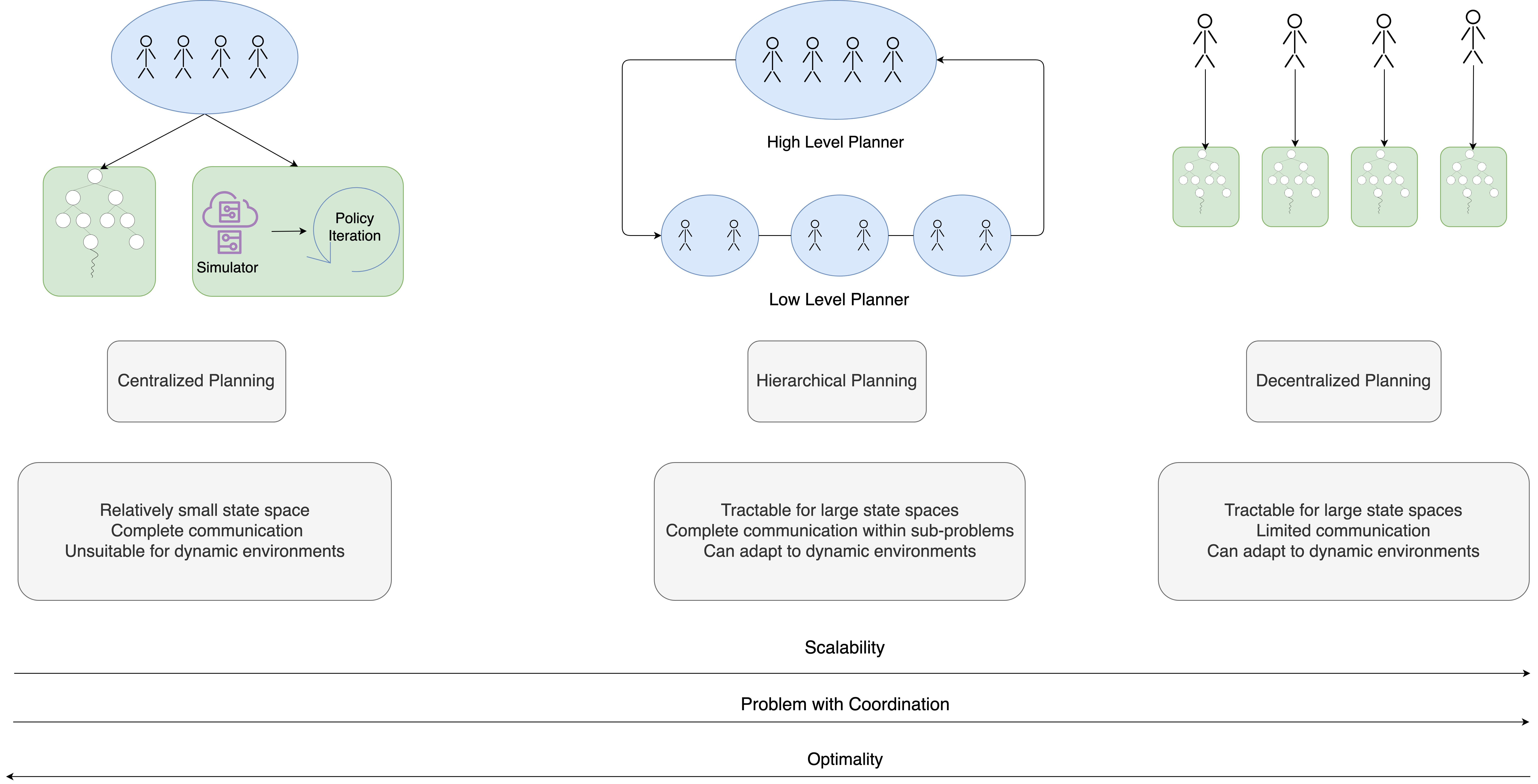}
\caption{The spectrum of decision theoretic models that can be used to tackle allocation and response in ERM pipelines. From left to right, scalability increases but coordination among agents and utility (in terms of optimality) decreases.}
\label{fig:planningSpectrum}
\end{center}
\end{figure*}

A problem that decision-theoretic methods pose for dynamic reallocation (and dispatch as well, depending on the size of the geographic area) is that the large state and action spaces can render standard approaches to be intractable. In such cases, approximation methods and intelligent heuristics can be used to leverage the structure of the specific problem to ensure scalability. One way is to use the canonical \textit{divide and conquer} approach by segregating the area under construction into sub-regions~\citep{pettet2020hierarchical}. Then, a smaller decision-theoretic problem can be solved for each sub-region. Another way to ensure scalability is to use decentralized planning~\citep{pettet2020algorithmic}, in which each agent plans for itself with locally available information. Such an approach can also be particularly useful for situations where communication systems are down (for example, in 2020, an explosion in Nashville, a city in the United States broke communication systems and hampered emergency response for four days~\citep{usaTodayNashville}). Both approaches have their own merits that we summarize in the next section (see Figure~\ref{fig:planningSpectrum}).

\subsection{Key Takeaways}
Allocation and response models are a crucial component of ERM pipelines, and a variety of algorithmic approaches have been used for allocating responders in anticipation of accidents. Recently, decision-theoretic approaches have been widely used for allocating and dynamically rebalancing the distribution of responders. We summarize the crucial takeaways from different decision-theoretic approaches that have been taken for emergency response in Figure~\ref{fig:planningSpectrum}. There are three major paradigms that have been used in this context --- centralized planning, hierarchical planning, and decentralized planning. Characteristics and features of each method are shown in Figure~\ref{fig:planningSpectrum}.

An extremely important area of focus in designing approaches for allocation and dispatch is the choice of the variable or metric that is optimized. Specifically, patient survival is a vital consideration that ambulances need to take into account while designing response models, since ambulances need to transport patients to medical facilities, which in turn increases the overall service time. This effect is naturally manifested in the choice of objectives and variables for allocation models. Despite this difference, there are high-level similarities in response modeling that apply to all emergency incidents (especially in reactive response). Models focusing on increasing coverage and reducing wait-times are common objectives that have been widely used in practice. Finally, the dynamic nature of urban areas must be factored in to the models as changes in the traffic and constructions can affect response times by ambulances.

Finally, we recommend model designers and practitioners to:
\begin{enumerate}[noitemsep]
    \item Be well-versed with the different objectives that have been used in response and allocation models, and carefully choose the one that suits the specific needs of the concerned area.
    \item Seek the help of domain experts (researchers, fire-fighters, etc.) to understand problems that responders face in the field. For example, the nearest ambulance might be heading in the opposite direction from the demand node on a highway, without the scope of making a turn. This makes it important to consider features that might not be intuitive to researchers.
    \item Seek to bridge the gap between theoretical models and realistic environmental constraints. For example, there is a rich body of work that makes the assumption that the environment in which ERM systems operate is static. While such assumptions simplify computational challenges, they might not truly capture the dynamics of actual ERM pipelines.
    \item Carefully evaluate the performance of predictive models and simulators before they are used for response and allocation.
    \item Consider prior work that uses well-crafted heuristics for scalability. 
    \item Smart and connected communities must plan in advance to account for events that might destroy communication infrastructure. Decentralized planning with locally available information can be promising in such situations.
\end{enumerate}
\section{Summary Discussion and Conclusion}\label{sec:future}

The field of designing emergency response pipelines has seen tremendous growth in the last few decades. Several factors have contributed to this growth including wider availability of data, the development of data-driven methodologies, increased cognizance, dependence and trust over algorithmic approaches by governments, and increase in computational power. However, there are still challenges in this field that need to be addressed. As we have pointed out, an EMS pipeline consists of an intricate combination of several components for its smooth functioning. There is a need for more research groups to: i) study EMS pipelines in their entirety, and consider the broader impact of their modular work on ERM systems, ii) consider and acknowledge the challenges and constraints that first responders face in the field, and iii) iteratively develop ERM tools by having first responder organizations in the loop. There are nuances that describe such needs throughout this paper. For example, an improved statistical fit for the prediction models does not necessarily mean an overall improvement for the ERM pipeline if the underlying model does not capture the true dynamics of incident occurrence. There is also a need for researchers to make their data and tools available to both the research community and ERM organizations. 

In a comprehensive review of statistical methods of crash prediction, \citeauthor{Lord2010}~\cite{Lord2010} pointed out that the wider availability of data is extremely promising for the field of crash prediction. This is particularly true now. Vast volumes of real-time data are now available from electric scooters, automobiles, and ambulances. There is also wider coverage of sensors like video-cameras throughout urban areas. This promise of increased availability of richer data holds true not only for incident data but also for data regarding covariates that potentially affect incident occurrence, like traffic congestion. The net result of an increased stream of data promises a finer understanding of the effect of covariates on incident occurrence. This benefit can be utilized by sharing data and algorithmic approaches between research groups and first responders.

Urban dynamics of accidents and crashes are continuously changing, and hold several opportunities for research. The increase in the number of automobiles and the arrival of autonomous vehicles in the markets across the globe presents the scope of re-evaluating existing models of crash occurrence and designing newer models that accommodate the changing landscape. \citeauthor{litman2017autonomous}~\cite{litman2017autonomous} lists the various additional planning constraints that need to be taken into account while developing transit systems that can accommodate autonomous vehicles, as well as additional causes for crashes, like software failure and increased overall travel volume. The potential risk factors caused by the interaction between autonomous and non-autonomous vehicles also pose challenges \cite{surveyAutonomous} and the need to design newer models of incident prediction.


Incident response also poses fresh challenges and opportunities. First, there is a need to combine the different metrics used in designing dispatch and allocation models. There are several interesting threads of research (cooperative coverage, survival metrics, gradual coverage decay, incorporating multiple resource types with different functionalities, etc.) that, to the best of our knowledge, have not been combined and evaluated together. Also, there has not been much focus on explicitly incorporating measures of patient survival directly in response models. We think that it is crucial that patient survival be studied in more detail and included as a part of objective functions for optimization approaches used in designing allocation and dispatch systems. 

A recent development in emergency response systems has been the computational ability of agents. Most modern ambulances and police vehicles are now equipped with laptops, which presents the scope of fast and decentralized decision-making, a particularly exciting area for multi-agent systems. Decentralized decision-making has been explored in the context of urban ERM systems~\cite{pettet2017incident}, but such approaches are probably more relevant for disaster scenarios like floods and hurricanes, where agents might lose connectivity to the central decision-making authority. Algorithmic approaches to aid the strategic redistribution of responders between incidents is extremely promising. While post-incident planning presents many technical challenges, such approaches rarely get implemented in the field. Inter-incident planning, on the other hand, respects the inherent challenges that emergency response faces. 

Finally, as smart and connected communities grow, it is crucial to ensure that resources are distributed and allocated in a manner that is equitable. As a result, equity of emergency response is also a concern as accessibility to emergency response can depend on the availability of financial resources and socio-economic backgrounds. For example, prior work has suggested bias in emergency services (specifically in drug administration) against minority communities~\cite{racialBiasEMS}. There is a need to first quantify fairness in the context of emergency response and then explicitly model such a notion in decision-making pipelines. As urban areas grow and witness a rise in population density, the need to design principled approaches to aid emergency response grows as well. This survey identifies how the field has evolved over the last few decades, with the view to aid researchers, policy-makers, and first-responders in designing better ERM pipelines.
\section{Acknowledgement}

The authors will like to thank the Center of Automotive Research at Stanford (CARS), the National Science Foundation (Grants: CNS-1640624, IIS-1814958, CNS-1818901) and the Tennessee Department of Transportation (TDOT) for funding the research. We would also like to express our gratitude towards the Nashville Fire Department (NFD), the Metropolitan Nashville Police Department (MNPD), the Nashville Metropolitan Information Technology Services (ITS) and the Tennessee Department of Transportation (TDOT) for invaluable feedback and domain expertise that made us understand the subtleties and intricate challenges of emergency response. Finally, we would like to thank Saideep Nannapaneni and Hemant Purohit for their feedback.


 \balance
\bibliography{references}

\end{document}